\documentclass[journal]{IEEEtran}
\usepackage{cite}

\usepackage{graphicx}  
\usepackage{amsmath}
\usepackage{amsthm}
\usepackage{array}
\usepackage{amsfonts}
\usepackage{bm}	
\usepackage[noend]{algpseudocode}
\usepackage{algorithmicx,algorithm}
\usepackage{subfig}
\theoremstyle{definition}
\newtheorem{thm}{Theorem}
\newtheorem{lamm}{Lemma}
\newtheorem{prop}{Proposition}
\usepackage{multirow}

\ifCLASSINFOpdf
\else
\fi
\begin{document}
%
\title{Robust Matrix Elastic Net based Canonical Correlation Analysis: An Effective Algorithm for Multi-View Unsupervised Learning}
%
%
%

\author{Peng-Bo~Zhang~and~
	Zhi-Xin~Yang,~\IEEEmembership{Member,~IEEE}
	\IEEEcompsocitemizethanks{\IEEEcompsocthanksitem Peng-Bo Zhang is with the Department
		of Industrial Engineering and Logistics Management, School of Engineering, Hong Kong University of Science and Technology, Kowloon, Hong Kong 999077, China. E-mail: pzhangag@connect.ust.hk (Corresponding author) 
		\IEEEcompsocthanksitem Zhi-Xin Yang is with the Department of Electromechanical Engineering, Faculty of Science and Technology, University of Macau, Macau 999078, China. E-mail:  zxyang@umac.mo}}

%
%

\markboth{}%
{Shell \MakeLowercase{\textit{et al.}}: Bare Demo of IEEEtran.cls for IEEE Journals}
%



\maketitle

\begin{abstract}
This paper presents a robust matrix elastic net based canonical correlation analysis (RMEN-CCA) for multiple view unsupervised learning problems, which emphasizes the combination of CCA and the robust matrix elastic net (RMEN) used as coupled feature selection. The RMEN-CCA leverages the strength of the RMEN to distill naturally meaningful features without any prior assumption and to measure effectively correlations between different 'views'. We can further employ directly the kernel trick to extend the RMEN-CCA to the kernel scenario with theoretical guarantees, which takes advantage of the kernel trick for highly complicated nonlinear feature learning. Rather than simply incorporating existing regularization minimization terms into CCA, this paper provides a new learning paradigm for CCA and is the first to derive a coupled feature selection based CCA algorithm that guarantees convergence. More significantly, for CCA, the newly-derived RMEN-CCA bridges the gap between measurement of relevance and coupled feature selection. Moreover, it is nontrivial to tackle directly the RMEN-CCA by previous optimization approaches derived from its sophisticated model architecture. Therefore, this paper further offers a bridge between a new optimization problem and an existing efficient iterative approach. As a consequence, the RMEN-CCA can overcome the limitation of CCA and address large-scale and streaming data problems. Experimental results on four popular competing datasets illustrate that the RMEN-CCA performs more effectively and efficiently than do state-of-the-art approaches.
\end{abstract}


%
\IEEEpeerreviewmaketitle

\section{Introduction}

\IEEEPARstart{W}{ith} the rapid development of the Internet, the amount of multiple view data available is extremely increasing in various domains. Since multiple representations of multi-view data are captured from various sources or different features spaces, the statistical properties of these representations, in general, are completely different. Therefore, learning and searching a consensus pattern in terms of effectiveness and efficiency from these multifarious representations is a persistent challenge. In order to address the above problem, many researchers have investigated widely many multi-view learning approaches \cite{yu2014high,zhu2016block,dhillon2011multi,wang2016comprehensive}. These existing multi-view learning approaches can be classified into three categories \cite{xu2013survey}: $1.$ co-training, $2.$ multiple kernel learning and $3.$ subspace learning. Our work focuses on the third one. 

Canonical Correlation Analysis (CCA) is a classical and powerful unsupervised learning approach for the multi-view learning problem. Its variants, such as Deep CCA (DCCA)\cite{andrew2013deep}, Randomized nonlinear CCA (FKCCA and NKCCA) \cite{lopez2014randomized}, Sparse CCA \cite{chu2013sparse, hardoon2011sparse} and Scalable CCA \cite{ma2015finding}, etc, have been thoroughly investigated. The basic idea of CCA is to find coupled linear projection matrices, and then model the potential connections between the two different 'views'. For this reason, CCAs have been widely applied to address multi-view learning problems.           

The two major challenges of CCA are: revealing the correlations across different sources in terms of effectiveness \cite{bach2002kernel,lopez2014randomized,wang2015large} and addressing the non-convex optimization problem of CCA efficiently \cite{ma2015finding,xie2015scale,wang2016efficient}. In this paper, we focus on the former. Over the past two decades, many studies have thoroughly investigated various kernel-based CCA approaches \cite{lopez2014randomized,wang2015large}. Although the performance of these kernel-based CCA approaches have been improved remarkably, these methods are not powerful enough to explore shared knowledge across different 'views'. In other words, these works ignore one critical property of the multi-view learning approach that is $\emph{coupled feature selection}$. Additionally, it is quite difficult to select a 'suitable' Mercer kernel in kernel-based approaches, which is a key factor for the success of kernel-based approaches. Generally speaking, the commonly-used Gaussian kernel cannot lead to the optimal performance. Therefore, these previous kernel-based CCA methods must consider the distribution of inputs and the corresponding application scenarios as prior knowledge. 

In this paper, we present a robust matrix elastic net based canonical correlation analysis (RMEN-CCA) without any prior knowledge. To the best of our knowledge, the RMEN-CCA is the first to incorporate the RMEN into CCA, thus emphasizing the combination of CCA with the coupled feature selection technique. In the RMEN, the $l_{21}$ norm allows the RMEN-CCA to capture joint sparse structure to distill relevant attributes from the data-embedding space, and simultaneously, the nuclear norm models the correlation between the projected samples via calculating a low-rank solution. More significantly, this paper provides a novel paradigm for CCA, which leverages the strength of coupled feature selection. Furthermore, in order for the RMEN-CCA to handle the highly sophisticated nonlinear relationship, this algorithm can perform directly in conjunction with kernel tricks, and we define this kernel scenario as KRMEN-CCA.

Moreover, we note that addressing CCA is a typical non-convex optimization problem because of its non-convex constraints, and thus it is nontrivial to solve it by the naive gradient descent directly. The AppGrad \cite{ma2015finding} is an efficient iterative algorithm for CCA. The crucial idea of the AppGrad is to guarantee that the domain of CCA exists in a convex region at each iteration. The AppGrad has two major advantages over the traditional eigenvector computation. First, the AppGrad can significantly decease computational and shortage complexity. Second, the online property of the AppGrad makes it efficient on handling huge datasets, whereas the eigenvector computation is prohibitive in this situation. Unfortunately, it is nontrivial to use directly the AppGrad approach to solve the RMEN-CCA because of the imposed RMEN. Therefore, based on the AppGrad baseline \cite{ma2015finding}, we derive a novel accelerated iterative method with proved convergence to address such non-convex optimization problem.

Furthermore, because of its highly flexible model architecture, the RMEN-CCA can be used as an intermediate structure in convolutional neural networks (CNNs) \cite{krizhevsky2012imagenet} for fine-tuning. This algorithm can also be applied to transfer learning \cite{pan2010survey} in order to address the problem of insufficient labeled data. However, these are beyond the scope of this paper. 

The \emph{contributions} of this paper are summarized as follows.
\begin{itemize}
	\item We present a novel robust matrix elastic net based canonical correlation analysis (RMEN-CCA) with theoretical guarantees and empirical proficiency. To the best of our knowledge, the RMEN-CCA is the first to incorporate the coupled feature selection into CCA, which improves generalization performance  by automatically distilling relevant and useful features without any prior knowledge. 
	\item  In the RMEN, the $l_{21}$ norm enforces redundancy and meaningless reduction, and meanwhile the nuclear norm yields a low-rank solution that encodes the correlation better. As a result, the RMEN-CCA takes advantage of this sparse plus low-rank structure, which brings benefits in terms of effectiveness and efficiency.
	\item The RMEN-CCA can directly leverage the powerful kernel trick to yield the kernel-based algorithm called KRMEN-CCA, which can effectively construct nonlinear approximations of the manifolds.
	\item It is nontrivial to address the proposed RMEN-CCA by existing approaches. Therefore, this paper bridges the gap between a novel non-convex optimization problem and the previous efficient iterative approach, which results that the RMEN-CCA can be applied to large-scale and online learning tasks.       
\end{itemize}

The remainder of the paper is organized as follows. Section \ref{sec:2} briefs the related work of feature selection techniques. Section \ref{sec:3} introduce the RMEN-CCA and its kernel version, and then an accelerated iterative optimization algorithm is derived for the RMEN-CCA. Section \ref{sec:4} analyzes theoretically the convergence of the RMEN-CCA. Section \ref{sec:exp} evaluates the RMEN-CCA on four popular datasets. Finally, conclusive remarks are provided in Section \ref{sec:6}.

\section{Related work of coupled feature selection}
\label{sec:2}
In recent years, the feature selection technique plays an important role in machine learning community. This technique is desired to extract numerous useful features and to eliminate redundancies, in order to construct a simple model architecture. To overcome the limitations of conventional feature selection approaches such as Lasso and Ridge, etc., a combined feature selection approach, termed matrix elastic net (MEN), has been successfully used for different learning algorithms \cite{gaiffas2011sharp,li2012error,zhen2017multi}. 

This paper defines a novel MEN termed RMEN, which uses the $l_{21}$ norm as joint feature selection instead of the Frobenius norm in the MEN. Our motivation is inspired by the LCFS algorithm \cite{wang2013learning} which incorporates this coupled feature selection into a linear regression method. However, in contrast to the LCFS, the RMEN-CCA confronts $\emph{two greater challenges}$: learning in an unsupervised fashion and addressing a non-convex optimization. Due to the high costs of labeling data manually, few labeled data may be available, even in this big data era. For this reason, deriving an efficient and effective approach without supervised information should be a hot topic and a promising research direction in machine learning community\cite{luo2017unsupervised,singh2017artificial,mirza2016generalizable}. Since there is no desired goal, the learning process of the RMEN-CCA is more difficult than the supervised learning algorithm LCFS. Therefore, how to distill numerous useful features and information plays an important role in the RMEN-CCA. On the other hand, in contrast to the LCFS that has the architecture of a linear system, the formula for the RMEN-CCA is more complicated because of its non-convex constrains. For this reason, this algorithm always fails to converge when using the naive gradient decent. Therefore, we must derive an effective optimization method to solve the novel RMEN-CCA.

The $l_{21}$ norm has been successfully used as joint feature selection in various domains. Argyriou et. al. \cite{argyriou2007multi} adopted the $l_{21}$ norm as the regularizer for multi-task feature learning tasks. Gu et. al. \cite{gu2011joint} derived a framework based on the $l_{21}$ norm for joint feature selection and subspace learning. Du et. al. \cite{du2015robust} presented a robust k-means approach based on the $l_{21}$ norm. We note that the $l_{21}$ norm is well suitable to our tasks, especially multi-view multi-tags problems. The $l_{21}$ norm has two major advantages over the Frobenius norm \cite{nie2010efficient}. The $l_{21}$ norm is much more robust to outliers than the Frobenius norm. More importantly, the $l_{21}$ norm considers joint sparse structure to choose relevant attributes across all samples, rather than being based on the importance of individual feature. In other words, the $l_{21}$ norm not only captures local useful information and potential relevant features, but also takes into account the manifold structure of feature space. As a result, we found that the RMEN-CCA not only outperforms the CCA with the MEN in multi-task learning problems, but it also applies to one-task applications better, which will be illustrated in Section \ref{sec:exp}. 

In addition to the $l_{21}$ norm in the RMEN, the nuclear norm (or trace norm) is a popular low-rank learning approach with widespread applications \cite{hsieh2014nuclear,grave2011trace,Gu2014Weighted,li2016constrained}.  The nuclear norm can yield a low rank solution, thus simplifying significantly the model architecture. It is a common perspective that only few elements of the instances contribute to a task. Moreover, different from previous work of the nuclear norm \cite{zhen2017multi}, the nuclear norm in the RMEN is implemented over all projected instances rather than only adjustable parameters. As a consequence, the nuclear norm can enforce the relevance of projected samples with connections.

\section{The robust matrix elastic net based canonical correlation analysis}

\label{sec:3} 

In this section, we brief the formulation of the RMEN-CCA in Section \ref{sec:3.1}, and then we extend the RMEN-CCA to the kernel version (KRMEN-CCA) in order to handle the nonlinear input-output relationships in Section \ref{sec:3.2}. Finally, we derive a new accelerated iterative algorithm to solve the RMEN-CCA in Section \ref{sec:3.3}.      
\subsection{The formulation of the RMEN-CCA}
\label{sec:3.1}
First of all, we briefly introduce the formulation of the classical CCA method and the robust matrix elastic net.

Given two variables $\bm{X} \in R^{d_1 \times n}$ and $\bm{Y} \in R^{d_2 \times n}$, the linear algebraic formulation of the classical CCA method \cite{golub1995canonical} is shown as follows.
\begin{align}
\label{eq1}
&\min_{\bm{U}, \bm{V}} \frac{1}{2n} \| \bm{X}^T\bm{U} - \bm{Y}^T\bm{V} \|^2_F \nonumber \\
&s.t. \quad \bm{U}^T\bm{X}\bm{X}^T\bm{U} = \bm{I}, \bm{V}^T\bm{Y}\bm{Y}^T\bm{V} = \bm{I}
\end{align}
where $\bm{U} \in R^{d_1 \times k}$ and $\bm{V} \in R^{d_2 \times k}$ are true canonical variables, $k$ is the number of the top canonical subspace, $\bm{I}$ is an identity matrix, and $\|\cdot\|_F$ is the Frobenius norm.

Given a matrix $\bm{Z} \in R^{n \times m}$, the $l_{21}$ norm $\|\bm{Z}\|_{21} = \sum_{i=1}^{n}\|\bm{Z}^{i}\|_2$ and the nuclear norm $\|\bm{Z}\|_{*} = \sum_{i=1}^{\min(m,n)}\sigma_i$, where $\bm{Z}^{i}$ is the $i$-th row of $\bm{Z}$ and $\sigma_i$ denotes the $i$-th singular value of the matrix $\textbf{Z}$.

Now we incorporate both the $l_{21}$ norm and the rank function into the classical CCA as follows.  
\begin{align}
\label{eq2}
&\min_{\bm{U}, \bm{V}} \frac{1}{2n} \| \bm{X}^T\bm{U} - \bm{Y}^T\bm{V} \|^2_F + \lambda_1 (\|\bm{U}\|_{21}+\|\bm{V}\|_{21}) \nonumber \\  &+\lambda_2(Rank[\bm{X}^T\bm{U} \quad \bm{Y}^T\bm{V}]) \nonumber \\
&s.t. \quad \bm{U}^T\bm{X}\bm{X}^T\bm{U} = \bm{I}, \bm{V}^T\bm{Y}\bm{Y}^T\bm{V} = \bm{I}
\end{align} 
where  $[\bm{X}^T\bm{U} \quad \bm{Y}^T\bm{V}]$ is a concatenated matrix. $\lambda_1$ and $\lambda_2$ are trading-off parameters. The former controls the $l_{21}$ norm for joint feature selection on two feature spaces simultaneously. The rank of the concatenated matrix is handled by the latter, that is, the larger $\lambda_2$ is, the lower the rank is.

It is clear that the rank function is noncontinuous, non-differentiable and non-convex. Hence, we use the nuclear norm instead of the rank function, which has been proven to be the tightest convex relaxation of the rank function \cite{recht2010guaranteed}. As a result, the (\ref{eq2}) is reformulated as
\begin{align}
\label{eq3}
&\min_{\bm{U}, \bm{V}} \frac{1}{2n} \| \bm{X}^T\bm{U} - \bm{Y}^T\bm{V} \|^2_F + \lambda_1 (\|\bm{U}\|_{21}+\|\bm{V}\|_{21}) \nonumber \\  &+\lambda_2(\|[\bm{X}^T\bm{U} \quad \bm{Y}^T\bm{V}]\|_{*}) \nonumber \\
&s.t. \quad \bm{U}^T\bm{X}\bm{X}^T\bm{U} = \bm{I}, \bm{V}^T\bm{Y}\bm{Y}^T\bm{V} = \bm{I}
\end{align} 
where $\|\cdot\|_{*}$ is the nuclear norm. We herein define the combination of the $l_{21}$ norm with the nuclear norm as a robust matrix elastic net (RMEN), and the (\ref{eq3}) is defined as the RMEN-CCA. 

Although the AppGrad \cite{ma2015finding} is an efficient iterative approach for the CCA, it is nontrivial to solve directly the (\ref{eq3}) by the AppGrad. To tackle such complicated non-convex problem, based on the AppGrad baseline, we derive an accelerated iterative approach with proved convergence. The details will be illustrated in Section \ref{sec:3.3}. 
\subsection{Kernel extension}
\label{sec:3.2}
In this section, we take advantage of the kernel trick to extend the proposed RMEN-CCA to nonlinear one capable of handling.

There exists a feature mapping: $\mathcal{X} \longrightarrow \phi_\mathcal{X}$ such that the following condition. For any two points $x_i,x_j \in \mathcal{X}$, we have $K_\mathcal{X}(x_i,x_j) = \langle \phi_\mathcal{X}(x_i),\phi_\mathcal{X}(x_j) \rangle $, where $K_\mathcal{X}(\cdot,\cdot)$ is a Mercer kernel and $\langle \cdot,\cdot \rangle$ is an inner product. We also deal with $\mathcal{Y}$ in the same fashion. Let $\Phi_\mathcal{X} = (\phi_\mathcal{X}(x_1), \dots, \phi_\mathcal{X}(x_n))^T$ and $\Phi_\mathcal{Y} = (\phi_\mathcal{Y}(y_1), \dots, \phi_\mathcal{Y}(y_n))^T$ be the new feature spaces. Following the \emph{Representer Theorem} \cite{dinuzzo2012representer}, the optimal solutions to the (\ref{eq3}) can be spanned by $\Phi_\mathcal{X}$ and $\Phi_\mathcal{Y}$. Therefore, we have $\bm{U} = \Phi_\mathcal{X}^T \bm{W}_\mathcal{X}$ and $\bm{V} = \Phi_\mathcal{Y}^T \bm{W}_\mathcal{Y}$, where $\bm{W}_\mathcal{X}, \bm{W}_\mathcal{Y} \in R^{n \times k}$ are the true canonical variables for the KRMEN-CCA. Consequently, the formula of the KRMEN-CCA is shown as
\begin{align}
\label{eq6}
\min_{\bm{W}_\mathcal{X}, \bm{W}_\mathcal{Y}}  \frac{1}{2n} \| \bm{K}_\mathcal{X}\bm{W}_\mathcal{X} - \bm{K}_\mathcal{Y}\bm{W}_\mathcal{Y} \|^2_F    +\lambda_1 \|\bm{W}_\mathcal{X}\|_{21}+ \nonumber\\ \lambda_1 \|\bm{W}_\mathcal{Y}\|_{21} + \lambda_2(\|[\bm{K}_\mathcal{X}\bm{W}_\mathcal{X} \quad \bm{K}_\mathcal{Y}\bm{W}_\mathcal{Y}]\|_{*}) \nonumber \\
s.t. \quad \bm{W}_\mathcal{X}^T\bm{K}_\mathcal{X}\bm{K}_\mathcal{X}\bm{W}_\mathcal{X} = \bm{I}, \bm{W}_\mathcal{Y}^T\bm{K}_\mathcal{Y}\bm{K}_\mathcal{Y}\bm{W}_\mathcal{Y} = \bm{I}
\end{align}

\subsection{An accelerated iterative algorithm for RMEN-CCA}
\label{sec:3.3}

It is quite difficult to address (\ref{eq3}) because of the two norms. Before presenting the novel accelerated iterative approach, we simplify the two norms.

Firstly, we simplify the $l_{21}$ norm. For the $l_{21}$ norm, there exists an unpredictable value when it is close to the origin \cite{he20122}. To overcome the above limitation of the $l_{21}$ norm, we need to define a function $\phi(\cdot)$ which satisfies all following conditions.
\begin{prop}\cite{he20122}
	\label{lamm1}
	Let $\phi(\cdot)$ be a function satisfying all following conditions,
	\begin{itemize}
		\item[1.] $ x \rightarrow \phi(x)$ is convex on $R$,
		\item[2.]  $x \rightarrow \phi(\sqrt{x})$  is concave on $R_+$,
		\item[3.] 	$\phi(x) = \phi(-x)$, $\forall x \in R$,
		\item[4.] $\phi(x)$ is $C^1$ on $R$,
		\item[5.] $\phi^{''}(0^+) > 0$, $\lim\limits_{x\rightarrow\inf} \phi(x)/x^2 = 0$.
	\end{itemize}
\end{prop}  

In this paper, we determine the function $\phi(x)$ as $\phi(x) = \sqrt{x^2+\zeta}$, where $\zeta$ is a small perturbation allowing the $l_{21}$ norm smoothness and differentiability. It is clear that the defined function $\phi(x)$ fulfills all conditions in Proposition \ref{lamm1}. 

After defining the function $\phi(\cdot)$, the following Lemma is helpful in solving this function in a half quadratic way \cite{nikolova2005analysis}.

\begin{lamm} \cite{he20122}
	\label{lamm4}
	Given $\|\bm{z}^i\|_2$, there is a a conjugate function
	$\varphi(\cdot)$, such that
	\begin{equation}
	\label{eq30}
	\phi(\|\bm{z}^i\|_2) = \inf_{a\in R} \{a\|\bm{z}^i\|^2_2+\varphi(a)\}
	\end{equation}
	where $a$ is determined by the minimizer function $\varphi(\cdot)$ w.r.t $\phi(\cdot)$ .
\end{lamm} 

Following the above Lemma, we can rewrite the (\ref{eq3}) in terms of traces as follows.
\begin{align}
\label{eq7}
&\min_{\bm{U}, \bm{V}} \frac{1}{2n} \| \bm{X}^T\bm{U} - \bm{Y}^T\bm{V} \|^2_F + \lambda_1 Tr(\bm{U}^T\bm{P}\bm{U}) \nonumber \\ &+ \lambda_1 Tr(\bm{V}^T\bm{Q}\bm{V}) +\lambda_2(\|[\bm{X}^T\bm{U} \quad \bm{Y}^T\bm{V}]\|_{*}) \nonumber \\
&s.t. \quad \bm{U}^T\bm{X}\bm{X}^T\bm{U} = \bm{I}, \bm{V}^T\bm{Y}\bm{Y}^T\bm{V} = \bm{I}
\end{align}
where $\bm{P} = diag(p^i)$ and $\bm{Q} = diag(q^j)$ $(i = 1, \dots d_1, j = 1, \dots, d_2)$. $p^i$ and $q^j$  can be calculated as
\begin{align}
\label{eq10}
p^i = \frac{1}{2\sqrt{\|\bm{u}^i\|^2_2+\zeta}} \nonumber \\
q^j = \frac{1}{2\sqrt{\|\bm{v}^j\|^2_2+\zeta}} 
\end{align} 
where $\bm{u}^i$ and $\bm{v}^j$ are the $i$-th row of matrix $\bm{U}$ and the $j$-th row of matrix $\bm{V}$, respectively, and $\zeta$ is a small smoothing term \cite{gorodnitsky1997sparse}. 

Subsequently, for the nuclear norm, the following Lemma presents a well-known variational formula.

\begin{lamm}\cite{grave2011trace,hsieh2014nuclear}
	\label{lamm2}
	Let $\bm{Z} \in R^{n \times m}$, The nuclear norm of $\bm{Z}$ is equivalent to
	\begin{equation}
	\label{eq31}
	\|\bm{Z}\|_* = \frac{1}{2}\inf_{\bm{S}\geq 0} Tr(\bm{Z}^T\bm{S}^{-1}\bm{Z} + \bm{S})
	\end{equation}  
	where $Tr(\bm{Z}^T\bm{S}^{-1}\bm{Z} + \bm{S})$ is a convex function (strictly convex when $\bm{Z}$ is invertible). If $\bm{Z}$ is invertible, the infinitum is obtained when $\bm{S} = \sqrt{(\bm{Z}\bm{Z}^T)}$. Otherwise, we can utilize $\sqrt{(\bm{Z}\bm{Z}^T)}$ from inside the cone of positive definite matrices to achieve arbitrarily close to the infimum. 
\end{lamm}

Following the above Lemma, substituting (\ref{eq31}) into (\ref{eq7}), we have
\begin{align}
\label{eq11}
\min_{\bm{U}, \bm{V}}&\min_{\bm{S} \geq 0}  \frac{1}{2n} \| \bm{X}^T\bm{U} - \bm{Y}^T\bm{V} \|^2_F + \nonumber \\ &\lambda_1 (Tr(\bm{U}^T\bm{P}\bm{U})+Tr(\bm{V}^T\bm{Q}\bm{V})) + \nonumber \\ &\lambda_2(Tr([\bm{X}^T\bm{U} \quad \bm{Y}^T\bm{V}]^T\bm{S}^{-1}[\bm{X}^T\bm{U} \quad \bm{Y}^T\bm{V}]) \nonumber \\ & +Tr(\bm{S})) \nonumber \\
s.t. \quad &\bm{U}^T\bm{X}\bm{X}^T\bm{U} = \bm{I}, \bm{V}^T\bm{Y}\bm{Y}^T\bm{V} = \bm{I}
\end{align}

Using the property of the nuclear norm, we can further simplify (\ref{eq11}) as
\begin{align}
\min_{\bm{U}, \bm{V}}& \min_{\bm{S} \geq 0} \frac{1}{2n} \| \bm{X}^T\bm{U} - \bm{Y}^T\bm{V} \|^2_F + \nonumber \\ & \lambda_1 (Tr(\bm{U}^T\bm{P}\bm{U})+Tr(\bm{V}^T\bm{Q}\bm{V})) + \nonumber \\ &\lambda_2(Tr(\bm{U}^T\bm{X}\bm{S}^{-1}\bm{X}^T\bm{U})+Tr(\bm{V}^T\bm{Y}\bm{S}^{-1}\bm{Y}^T\bm{V}) \nonumber \\ &+Tr(\bm{S})) \nonumber \\
s.t. \quad &\bm{U}^T\bm{X}\bm{X}^T\bm{U} = \bm{I}, \bm{V}^T\bm{Y}\bm{Y}^T\bm{V} = \bm{I}
\end{align}

Likewise to the $l_{21}$ norm \cite{gorodnitsky1997sparse}, we also impose an additional term $\zeta Tr(\bm{S}^{-1})$ for guaranteeing convergence  \cite{grave2011trace}. As a result, the infimum over $\bm{S}$ is achieved when
\begin{equation}
\bm{S} = \sqrt{\bm{X}^T\bm{U}\bm{U}^T\bm{X}+\bm{Y}^T\bm{V}\bm{V}^T\bm{Y}+\zeta\bm{I}}
\end{equation}

Finally, we have the aftermost optimization formula over $\bm{U}$ and $\bm{V}$ for given $\bm{S}$ as
\begin{align}
\label{eq14}
&\min_{\bm{U}, \bm{V}} \frac{1}{2n} \| \bm{X}^T\bm{U} - \bm{Y}^T\bm{V} \|^2_F + \nonumber \\ & \lambda_1 (Tr(\bm{U}^T\bm{P}\bm{U})+Tr(\bm{V}^T\bm{Q}\bm{V})) + \nonumber \\ &\lambda_2(Tr(\bm{U}^T\bm{X}\bm{S}^{-1}\bm{X}^T\bm{U})+Tr(\bm{V}^T\bm{Y}\bm{S}^{-1}\bm{Y}^T\bm{V})) \nonumber \\
&s.t. \quad \bm{U}^T\bm{X}\bm{X}^T\bm{U} = \bm{I}, \bm{V}^T\bm{Y}\bm{Y}^T\bm{V} = \bm{I}
\end{align}

To find the solutions to (\ref{eq14}), according to the AppGrad \cite{ma2015finding}, we derive an accelerated iterative approach. The new iterative approach has similar idea to the AppGrad. We also introduce an unnormalized pair $(\tilde{\bm{U}},\tilde{\bm{V}})$, which are updated via gradient descent with the momentum at each iteration. Subsequently, at this iteration, the true canonical pair  $(\bm{U}, \bm{V})$ are updated by the resulted unnormalized pair. However, different from the AppGrad, the novel iterative approach needs to calculate the intermediate variables $\bm{S}$, $\bm{P}$ and $\bm{Q}$ by the true canonical pair at each iteration. After that, these resulted intermediate variables are fed into the following updating step to calculate the unnormalized pair (see the loop in Algorithm \ref{alg:1}). 

At each iteration, we take the partial derivative of the objective function with respect to $\tilde{\bm{U}}$, and then update the $\tilde{\bm{U}}$ of the past time step to the current $\tilde{\bm{U}}$ using the momentum $\Delta_U$.
\begin{align}
\label{eq15}
\partial \tilde{\bm{U}} = \frac{1}{n}\bm{X}(\bm{X}^T\tilde{\bm{U}}-\bm{Y}^T\bm{V}) + \lambda_1 \bm{P}\tilde{\bm{U}} + \lambda_2 \bm{X} \bm{S}^{-1} \bm{X}^T \tilde{\bm{U}} \nonumber \\
\Delta_U = \gamma \Delta_U - \eta \partial \tilde{\bm{U}} \nonumber \\
\tilde{\bm{U}} = \tilde{\bm{U}} + \Delta_U
\end{align}
where $\gamma$ is a momentum coefficient that controls the momentum and $\eta$ is a learning rate.

After updating $\tilde{\bm{U}}$, we then calculate the true canonical variable $\bm{U}$ such that $\bm{U}^T\bm{X}\bm{X}^T\bm{U} = \bm{I}$. 
\begin{align}
\label{eq16}
SVD: \bm{L}_x^T\bm{\Sigma}_x\bm{L}_x = \tilde{\bm{U}}^T(\frac{1}{n}\bm{X}\bm{X}^T)\tilde{\bm{U}} \nonumber \\
\bm{U} = \tilde{\bm{U}}\bm{L}_x^T\bm{\Sigma}_x^{-\frac{1}{2}} \bm{L}_x
\end{align}
where $\bm{L}_x$ and $\bm{\Sigma}_x$ is a unitary matrix and a rectangular diagonal matrix, respectively.

Likewise to (\ref{eq15}) and (\ref{eq16}), we can calculate the other true canonical variable $\bm{V}$ in the same fashion.
\begin{align}
\label{eq17}
\partial \tilde{\bm{V}} =  \frac{1}{n}\bm{Y}(\bm{Y}^T\tilde{\bm{V}}-\bm{X}^T\bm{U}) + \lambda_1 \bm{Q}\tilde{\bm{V}} + \lambda_2 \bm{Y} \bm{S}^{-1} \bm{Y}^T \tilde{\bm{V}} \nonumber \\
\Delta_V = \gamma \Delta_V - \eta \partial \tilde{\bm{V}} \nonumber \\
\tilde{\bm{V}} = \tilde{\bm{V}} + \Delta_V
\end{align}
\begin{align}
\label{eq18}
SVD: \bm{L}_y^T\bm{\Sigma}_y\bm{L}_y = \tilde{\bm{V}}^T(\frac{1}{n}\bm{Y}\bm{Y}^T)\tilde{\bm{V}} \nonumber \\
\bm{V} = \tilde{\bm{V}}\bm{L}_y^T\bm{\Sigma}_y^{-\frac{1}{2}} \bm{L}_y
\end{align}
To sum up, the pseudo-code of the RMEN-CCA is summarized in Algorithm \ref{alg:1}.
\begin{algorithm}[htp]
	\caption{An Iterative Algorithm for RMEN-CCA }
	\label{alg:1}
	\hspace*{0.02in} {\bf Input:}
	Training data $\bm{X} \in R^{d_1 \times n} $ $\bm{Y} \in R^{d_2 \times n}$; the learning rate $\eta$; the trading-off parameters $\lambda_1,\lambda_2$; the momentum coefficient $\gamma$. \\
	\hspace*{0.02in} {\bf Output:} 
	The true canonical pair $(\bm{U}, \bm{V})$
	\begin{algorithmic}[1]
		\State Initialize $(\bm{U},\bm{V})$ from the standard Gaussian distribution, as well as set $(\tilde{\bm{U}},\tilde{\bm{V}})$ and the momentum $(\Delta_u, \Delta_v)$ as zero matrices;
		\While{not convergence} 
		\State Calculate $\bm{S}$ by singular value decomposition (SVD): $\bm{\Phi} \bm{S} \bm{\Phi}^T = (\bm{X}^T \bm{U}\bm{U}^T \bm{X}+ \bm{Y}^T \bm{V}\bm{V}^T \bm{Y}) $; 
		\State Calculate $\bm{S}^{-1} = \bm{\Phi}(\bm{S}+\zeta \bm{I})^{-\frac{1}{2}}\bm{\Phi}^T$;
		\State Calculate each diagonal element of $\bm{P}$ and $\bm{Q}$ using  (\ref{eq10}); 
		\State Update $\tilde{\bm{U}}$ and $\bm{U}$ in (\ref{eq15}) and (\ref{eq16}), respectively;
		\State Update $\tilde{\bm{V}}$ and $\bm{V}$ in  (\ref{eq17}) and (\ref{eq18}), respectively;
		\EndWhile
		\State \Return $(\bm{U}, \bm{V})$
	\end{algorithmic}
\end{algorithm}

Moreover, we can address directly the KMEN-CCA by the Algorithm \ref{alg:1}, when using $(\bm{K}_\mathcal{X}, \bm{K}_\mathcal{Y})$ and $(\bm{W}_\mathcal{X},\bm{W}_\mathcal{Y})$ to replace $(\bm{X},\bm{Y})$ and $(\bm{U},\bm{V})$, respectively. 

In order to boost further the generalization performance, we also present a stochastic iterative algorithm for RMEN-CCA, as shown in Algorithm \ref{alg:2}. 
\begin{algorithm}[htp]
	\caption{A Stochastic Iterative Algorithm for RMEN-CCA }
	\label{alg:2}
	\hspace*{0.02in} {\bf Input:}
	Training data $\bm{X} \in R^{d_1 \times n} $ $\bm{Y} \in R^{d_2 \times n}$; the learning rate $\eta$; the trading-off parameters $\lambda_1,\lambda_2$; the momentum coefficient $\gamma$. \\
	\hspace*{0.02in} {\bf Output:} 
	The true canonical pair $(\bm{U}, \bm{V})$
	\begin{algorithmic}[1]
		\State Initialize $(\bm{U},\bm{V})$ from the standard Gaussian distribution, as well as set $(\tilde{\bm{U}},\tilde{\bm{V}})$ and the momentum $(\Delta_u, \Delta_v)$ as zero matrices;
		\While{not convergence} 
		\State Randomly select a subset $\Omega$ of training samples, where we have $\bm{X}_\Omega \in R^ {d_1 \times m}$ and $\bm{Y}_\Omega \in R^{d_2 \times m}$ $(m \ll n)$; 
		\State Calculate $\bm{S}$ by singular value decomposition (SVD): $\bm{\Phi} \bm{S} \bm{\Phi}^T = (\bm{X}^T_\Omega \bm{U}\bm{U}^T \bm{X}_\Omega+ \bm{Y}^T_\Omega \bm{V}\bm{V}^T \bm{Y}_\Omega) $; 
		\State Calculate $\bm{S}^{-1} = \bm{\Phi}(\bm{S}+\zeta \bm{I})^{-\frac{1}{2}}\bm{\Phi}^T$;
		\State Calculate each diagonal element of $\bm{P}$ and $\bm{Q}$ using (\ref{eq10}); 
		\State Update $\tilde{\bm{U}}$ and $\bm{U}$ in (\ref{eq15}) and (\ref{eq16}), respectively;
		\State Update $\tilde{\bm{V}}$ and $\bm{V}$ in (\ref{eq17}) and (\ref{eq18}), respectively;
		\EndWhile
		\State \Return $(\bm{U}, \bm{V})$
	\end{algorithmic}
\end{algorithm}

\section{Convergence Analysis of the RMEN-CCA}
\label{sec:4}

The consistently fast convergence can illustrate the great practicality and efficiency of an algorithm. For this reason, it is worthwhile to discuss the convergence of the RMEN-CCA. In this section, we theoretically analyze the convergence of the RMEN-CCA in Theorem \ref{thm1}. Additionally, empirical studies in Section \ref{sec:exp} further illustrate the quick convergence of the RMEN-CCA in some real-world tasks. In order to prove Theorem \ref{thm1}, we firstly show the following Lemma which is helpful in proving Theorem \ref{thm1}. Lemma \ref{lamm3} not only reveals the relationship between the optimal true canonical pair and its unnormalized counterpart, but it also interprets the novel iterative approach as approximate gradient scheme for addressing the RMEN-CCA. 
  
\begin{lamm}
	\label{lamm3}
	Let $(\bm{U}^*,\bm{V}^*)$ and $(\tilde{\bm{U}}^*,\tilde{\bm{V}}^*)$ be the optimal true canonical pair and the optimal unnormalized pair, respectively. Then, we have $(\tilde{\bm{U}}^*,\tilde{\bm{V}}^*) = (\bm{\Lambda}\bm{U}^*,\bm{\Lambda}\bm{V}^*)$, where $\bm{\Lambda}$ is the canonical correlation diagonal matrix.
	
	\begin{proof}
		Let 
		\begin{align}
		\label{eq:1}
		\tilde{\bm{U}}^* &= \arg\min_{\bm{U}} \frac{1}{2n} \|\bm{X}^T\bm{U}-\bm{Y}^T\bm{V}^*\|^2_F + \nonumber \\ &\lambda_1 (Tr(\bm{U}^T\bm{P}\bm{U}))  + \lambda_2(Tr(\bm{U}^T\bm{X}\bm{S}^{-1}\bm{X}^T \bm{U})) \\
		\label{eq:3}
		\tilde{\bm{V}}^* &=  \arg\min_{\bm{V}} \frac{1}{2n} \|\bm{Y}^T\bm{V}-\bm{X}^T\bm{U}^*\|^2_F + \nonumber \\ &\lambda_1 (Tr(\bm{V}^T\bm{Q}\bm{V}))  + \lambda_2(Tr(\bm{V}^T\bm{Y}\bm{S}^{-1}\bm{Y}^T\bm{V}))
		\end{align}
		
		We only proof that the result $\tilde{\bm{U}}^* = \bm{\Lambda}\bm{U}^*$ holds, and then a similar result also holds for the other variable. 
		
		Following the optimality condition, we have 
		\begin{align}
		\label{eq:2}
		\frac{1}{n}\bm{X}\bm{X}^T\tilde{\bm{U}}^* = \frac{1}{n}\bm{X}\bm{Y}^T\bm{V}^*
		\end{align} 
		
		Using the Lemma 1.1 and 2.2 in \cite{ma2015finding}, we can directly obtain the desired result as follows.
		\begin{align}
		\tilde{\bm{U}}^* = \bm{\Lambda}\bm{U}^*
		\end{align} 
		
		Similar argument gives $\tilde{\bm{V}}^* = \bm{\Lambda}\bm{V}^*$.
		
		We complete the proof.
	\end{proof}
\end{lamm} 

Following the above Lemma, Theorem \ref{thm1} illustrates the main convergence result of the RMEN-CCA.
\begin{thm}
	\label{thm1}
	A sequence of the leading canonical pair $(\bm{U},\bm{V})$ in the RMEN-CCA converge to the optimal canonical pair $(\bm{U}^*,\bm{V}^*)$.    
	\begin{proof}
		The Lemma \ref{lamm3} shows the relationship between the two canonical pairs. Therefore, now we only consider the unnormalized canonical variables $(\tilde{\bm{U}}^*,\tilde{\bm{V}}^*)$. The newly-derived iterative approach is an alternating optimization, that is, at each iteration, given a fixed canonical variable, we update the other canonical variable. Therefore, we can reformulate the optimization form of the RMEN-CCA at the $t$-th iteration as
		\begin{align}
		\label{eq:18}
		\min_{\tilde{\bm{U}}}& \frac{1}{2n} \| \bm{X}^T\tilde{\bm{U}} - \bm{Y}^T\bm{V}_{t} \|^2_F + \lambda_1 (Tr(\tilde{\bm{U}}^T\bm{P}_t\tilde{\bm{U}}) + \nonumber \\ &\lambda_2(Tr(\tilde{\bm{U}}^T\bm{X}\bm{S}^{-1}_t\bm{X}^T\tilde{\bm{U}})) \\
		\label{eq19}
		\min_{\tilde{\bm{V}}}& \frac{1}{2n} \| \bm{Y}^T\tilde{\bm{V}} - \bm{X}^T\bm{U}_{t} \|^2_F + \lambda_1 (Tr(\tilde{\bm{V}}^T\bm{Q}_t\tilde{\bm{V}})) + \nonumber \\ &\lambda_2(Tr(\tilde{\bm{V}}^T\bm{Y}\bm{S}^{-1}_t\bm{Y}^T\tilde{\bm{V}}))
		\end{align}
		
		Without loss of generality, we only consider the (\ref{eq:18}), and the (\ref{eq19}) has the similar argument.
		
		According to \cite{nie2010efficient}, we illustrate that the second term monotonically decreases at each iteration. At the $t$-th iteration, we have
		\begin{align}
		\label{eq20}
		\tilde{\bm{U}}_{t+1} = \arg\min_{\tilde{\bm{U}}} Tr(\tilde{\bm{U}}^T\bm{P}_t\tilde{\bm{U}})
		\end{align}
		
		The Eq. \ref{eq20} indicates that
		\begin{align}
		\label{eq21}
		Tr(\tilde{\bm{U}}_{t+1}^T\bm{P}_{t+1}\tilde{\bm{U}}_{t+1}) \leq Tr(\tilde{\bm{U}}_t^T\bm{P}_t\tilde{\bm{U}}_t)
		\end{align}
		
		The  (\ref{eq21}) can be written as    
		\begin{align}
		\label{eq22}
		\sum_{i=1}^{d_1}\frac{\|\tilde{\bm{u}}_{t+1}^i\|^2_2}{2\|\tilde{\bm{u}}_{t+1}^i\|_2} \leq \sum_{i=1}^{d_1}\frac{\|\tilde{\bm{u}}_{t}^i\|^2_2}{2\|\tilde{\bm{u}}_{t}^i\|_2}
		\end{align}  
		
		Following an obvious inequality $(\sqrt{\kappa_1} - \sqrt{\kappa_2})^2 \geq 0$, for each $i$, we have 
		\begin{align}
		\label{eq23}
		\|\tilde{\bm{u}}_{t+1}^i\|_2- \frac{\|\tilde{\bm{u}}_{t+1}^i\|^2_2}{2\|\tilde{\bm{u}}_{t+1}^i\|_2} \leq 
		\|\tilde{\bm{u}}_{t}^i\|_2-\frac{\|\tilde{\bm{u}}_{t}^i\|^2_2}{2\|\tilde{\bm{u}}_{t}^i\|_2}
		\end{align}
		
		Summing over all above inequalities, we have  
		\begin{align}
		\label{eq24}
		\sum_{i=1}^{d_1}(\|\tilde{\bm{u}}_{t+1}^i\|_2- \frac{\|\tilde{\bm{u}}_{t+1}^i\|^2_2}{2\|\tilde{\bm{u}}_{t+1}^i\|_2}) \leq 
		\sum_{i=1}^{d_1}(\|\tilde{\bm{u}}_{t}^i\|_2-\frac{\|\tilde{\bm{u}}_{t}^i\|^2_2}{2\|\tilde{\bm{u}}_{t}^i\|_2})
		\end{align}
		
		Combining (\ref{eq22}) and (\ref{eq24}), we have	
		\begin{align}
		\sum_{i=1}^{d_1}\|\tilde{\bm{u}}_{t+1}^i\|_2 \leq \sum_{i=1}^{d_1}\|\tilde{\bm{u}}_{t}^i\|_2
		\end{align}
		
		That is to say,
		\begin{align}
		\|\tilde{\bm{U}}_{t+1}\|_{21} \leq \|\tilde{\bm{U}}_t\|_{21} 
		\end{align}
		which is the desired result.  	
		
		Moreover, the Lemma \ref{lamm2} indicates that the third term $Tr(\tilde{\bm{U}}^T\bm{X}\bm{S}_t^{-1}\bm{X}^T\tilde{\bm{U}})$ converges to the optimum $\tilde{\bm{U}}^*$. Since the nuclear norm is the infimum over $\bm{S}$.
		
		Now we revisit the (\ref{eq:18}). This objective function consists of three norms. And thus, the objective function is convex and the value is not less than 0. When using the gradient descend, the objective function is still moving towards the negative direction of the gradient. Therefore, the unnormalized canonical variables $(\tilde{\bm{U}}_t,\tilde{\bm{V}}_t)$ are monotonically decreasing as $t \rightarrow +\infty$, and converge to the optimal unnormalized canonical variables $(\tilde{\bm{U}}^*,\tilde{\bm{V}}^*)$. 	
		
		Furthermore, we revisit the original optimization problem of the RMEN-CCA. Following the Lemma \ref{lamm3}, we have the following statement that when the unnormalized pair $(\tilde{\bm{U}}_t,\tilde{\bm{V}}_t)$ converge to $(\tilde{\bm{U}}^*,\tilde{\bm{V}}^*)$ as $t \rightarrow +\infty$, the true canonical pair $(\bm{U}_t,\bm{V}_t)$ also converge to $(\bm{U}^*,\bm{V}^*)$. We obtain the desired result and complete the proof. 
	\end{proof}
\end{thm} 

\section{Experimental results and discussion}
\label{sec:exp}
\begin{figure*}[htpb]
	\centering
	\includegraphics[width=6.8in]{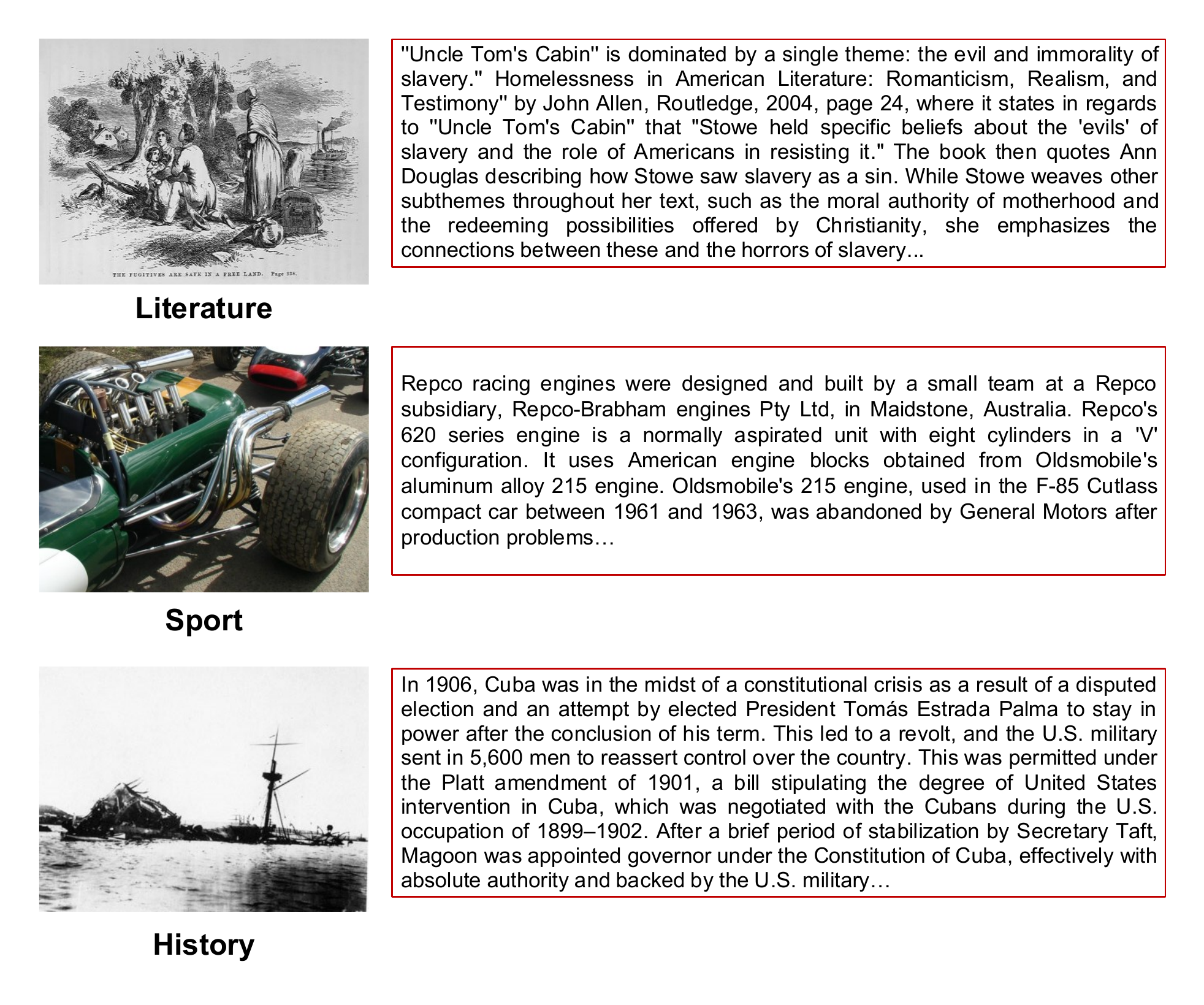}
	\DeclareGraphicsExtensions.
	\caption{Three examples on Wiki: image (left) and its corresponding describing article (right).}
	\label{fig:wiki}
\end{figure*}

\begin{figure*}[htpb]
	\centering
	\includegraphics[width=6.8in]{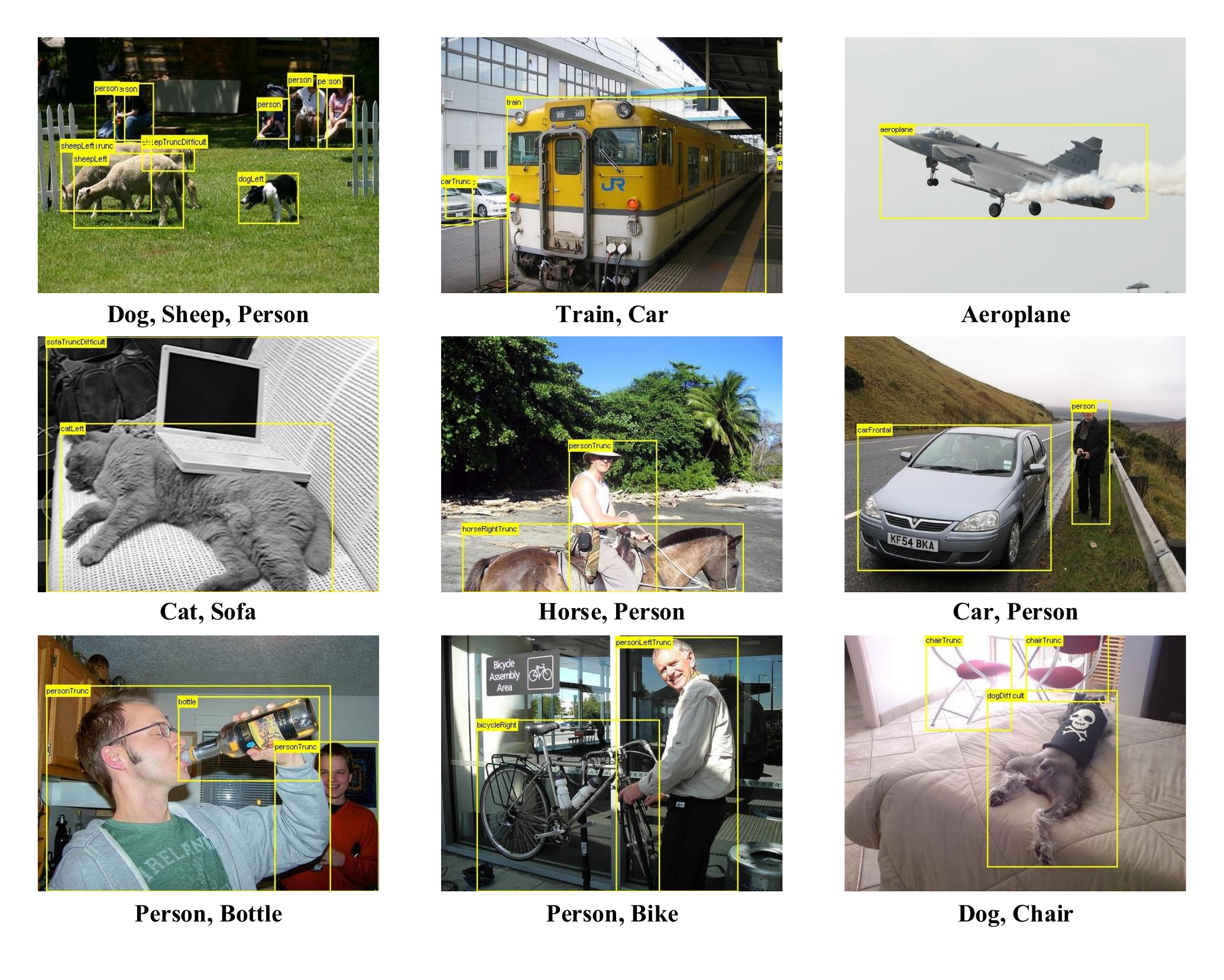}
	\DeclareGraphicsExtensions.
	\caption{Some examples on Pascal VOC 2007: image and its multiple tags.}
	\label{fig:poc}
\end{figure*}

In this section, we conduct several experiments on four popular datasets to evaluate the RMEN-CCA, including MNIST \cite{lecun1998gradient}, Wiki \cite{rasiwasia2010new}, Pascal VOC 2007 \cite{pascal-voc-2007} and URL Reputation \cite{ma2009identifying}. We randomly select $20\%$ training data as the validation data for each dataset, and the hyper-parameters are tuned on the validation set. The description of each dataset is shown as follows, and the statistics of these four datasets are summarized in Table \ref{table1}. 
\begin{itemize}
	\item The MNIST is a handwritten digit recognition dataset, in which each image is separated into the left and right halves. 
	\item The Wiki is a popular image-text matching dataset, which was used in \cite{wang2013learning,rasiwasia2010new,pereira2014role}. The Wiki dataset assembles from Wikipedia's articles, which consists of 2866 image-text pair (2173 training data and 693 test data). The features of images and texts are extracted by a 128-dim SIFT and a 10-dim latent Dirichlet allocation algorithm, respectively. Some examples are shown in Figure \ref{fig:wiki}. 
	\item The Pascal VOC 2007 is a challenging and realistic image and its multi-tags matching testbed. In order to carry out more experiments, three image features and three tag features are extracted respectively by \cite{hwang2010accounting}, including Gist, HSV color histograms and bag-of-visual-words (BoW) histograms, as well as word frequency (Wc), relative tag rank (Rel) and absolute tag rank (Abs). Some examples are shown in Figure \ref{fig:poc}.  
	\item The URL Reputation is a large-scale dataset for online learning algorithms. Due to limitations of computational resource, we only choose 1 Million samples and 50 attributes. 	
\end{itemize}

\begin{table*}[h]
	\renewcommand{\arraystretch}{1.3}
	\caption{The statistics of the four datasets}
	\label{table1}
	\centering
	\begin{tabular}{|c|c|c|c|c|c|}
		\hline
		Problems & Description& Training set& Test set & Dim X & Dim Y \\
		\hline
		MNIST &	Left and Right Halves of Images & 60,000& 10,000& 392& 392\\
		\hline
		Wiki & Image-Text Pairs  & 2,173 & 693 & 128 & 10 \\
		\hline 
		\multirow{3}{*}{Pascal VOC 2007} & \multirow{3}{*}{Image and Its Multi-Labels}&  \multirow{3}{*}{5,011} & \multirow{3}{*}{4,952} & 512 (Gist) & 399 (Wc) \\
		& & & & 200 (Bow) & 399 (Rel) \\
		& & & &  64 (Hsv) & 399 (Abs) \\
		\hline
		URL Reputation & Host and Lexical based Features& 1,000,000 & 20,000 & 50 & 50\\
		\hline
	\end{tabular}
\end{table*} 

Moreover, we measure the test results by the commonly-used Pearson product moment correlation coefficients (PCC). The PCC is defined as $PCC = \frac{cov(\bm{A},\bm{B})}{\sigma_A\sigma_B}$, where $\bm{A} = \bm{X}^T_{test}\bm{U}$ and $\bm{B} = \bm{Y}^T_{test}\bm{V}$ are two projected test samples, $cov$ is the covariance, as well as $\sigma_A$ and $\sigma_B$ are the standard deviation of $\bm{A}$ and $\bm{B}$, respectively. The range of PPC is from 100$(\%)$ to 0, in which 100 denotes complete correlation and 0 denotes no correlation. Ten trials are conducted for each algorithm, and we report the average PCC results. In the experiments, we found the number of dimensional canonical subspace $k$ is almost no effect on the performance. Therefore, except for MNIST using $k = 50$ , we calculate the top 5 $(k=5)$ dimensional canonical subspace for other three datasets. All simulations are carried out in a Matlab 2015b environment running in a PC machine with Inter(R) Core(TM) i7-6700HD 2.60 GHZ CPU, NVIDIA GTX 960M GPU and 8 GB of RAM.

\subsection{Hyper-parameters selection}
In the RMEN-CCA, we initialize the true canonical pair $(\bm{U},\bm{V})$ by drawing i.i.d samples from the standard Gaussian distribution, and set the unnormalized pair $(\tilde{\bm{U}},\tilde{\bm{V}})$ and the momentum $(\Delta_u, \Delta_v)$ as zero matrices. We empirically found that the RMEN-CCA is insensitive to the learning rate $\eta$ and the trade-off coefficients $\lambda_1$, $\lambda_2$. When the value of $\lambda_1$ is 10 times that of $\lambda_2$, the RMEN-CCA can achieve the best performance. Hence, we set $\eta = 0.005$, $\lambda_1 =0.01$ and $\lambda_2 = 0.001$ on all four datasets. According to \cite{qian1999momentum}, the momentum coefficient $\gamma$ is set as 0.9. For the stochastic iterative algorithm, a small part of the training data is held out. We use stochastic iterative approach for the RMEN-CCA in all experiments.  

We choose the commonly-used Gaussian kernel for all kernel based methods, and the kernel parameter is chosen empirically from $\{2^{-10}, 2^{-9}, \dots, 2^9, 2^{10}\}$ on the validation set. In addition, the number of random projections of both  FKCCA and NKCCA \cite{lopez2014randomized} are chosen from $\{1,000, 2,000, \dots, 5,000\}$, as well as that of approximate KCCA (KNOI) \cite{wang2015large} is from $\{10,000, 20,000, \dots, 100,000\}$. Other user-defined parameters are determined empirically, and we pick the one having the best performance. Due to space limitations, we omit the procedure of selecting hyper-parameters in this paper. 
\subsection{Convergence}

\begin{figure*}[ht]
	\renewcommand{\arraystretch}{1.3}
	\centering
	\subfloat[MNIST]{
		\centering
		\label{fig:1a}
		\includegraphics[width=3.3in]{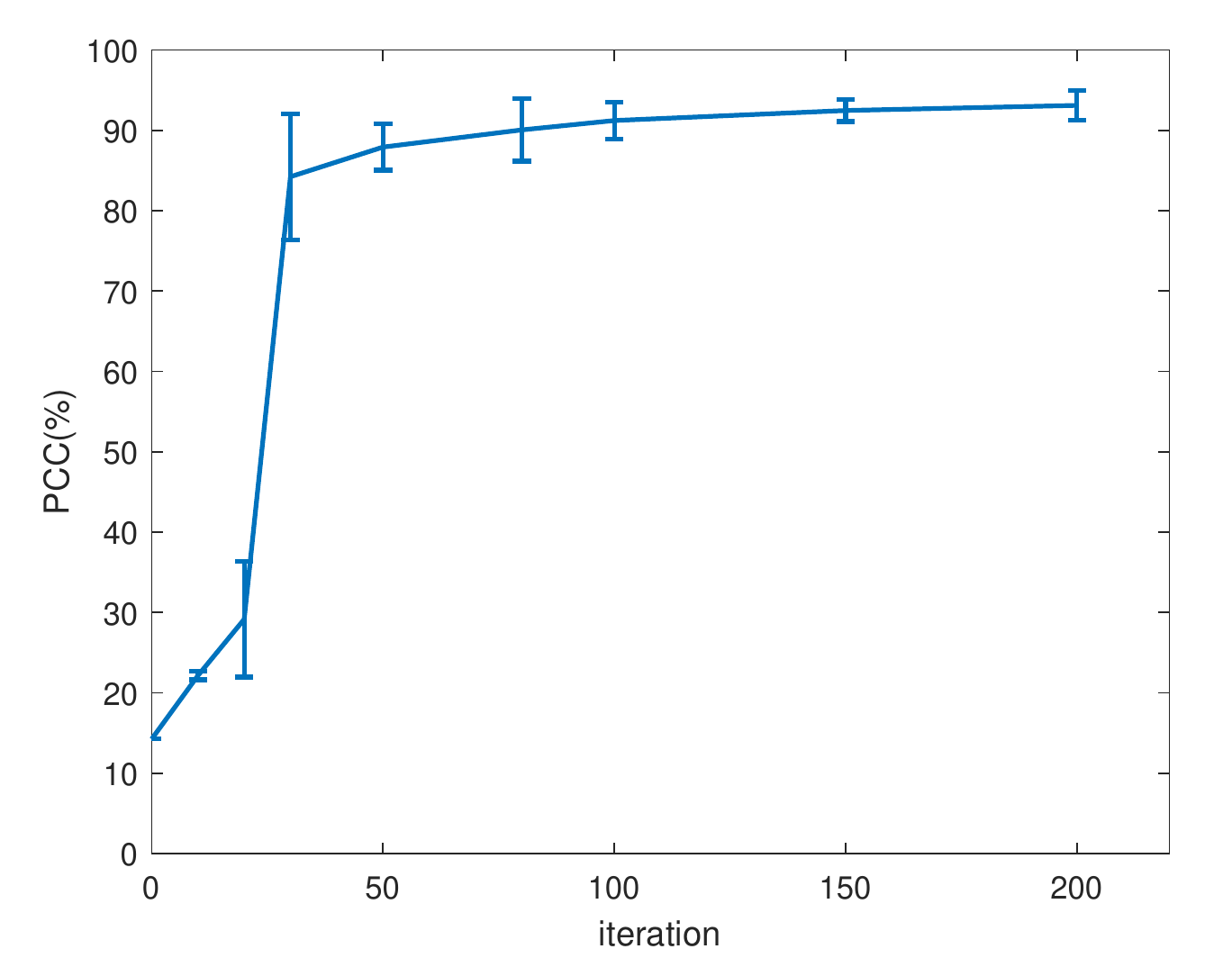}}
	\subfloat[Wiki]{
		\centering
		\label{fig:1b}
		\includegraphics[width=3.3in]{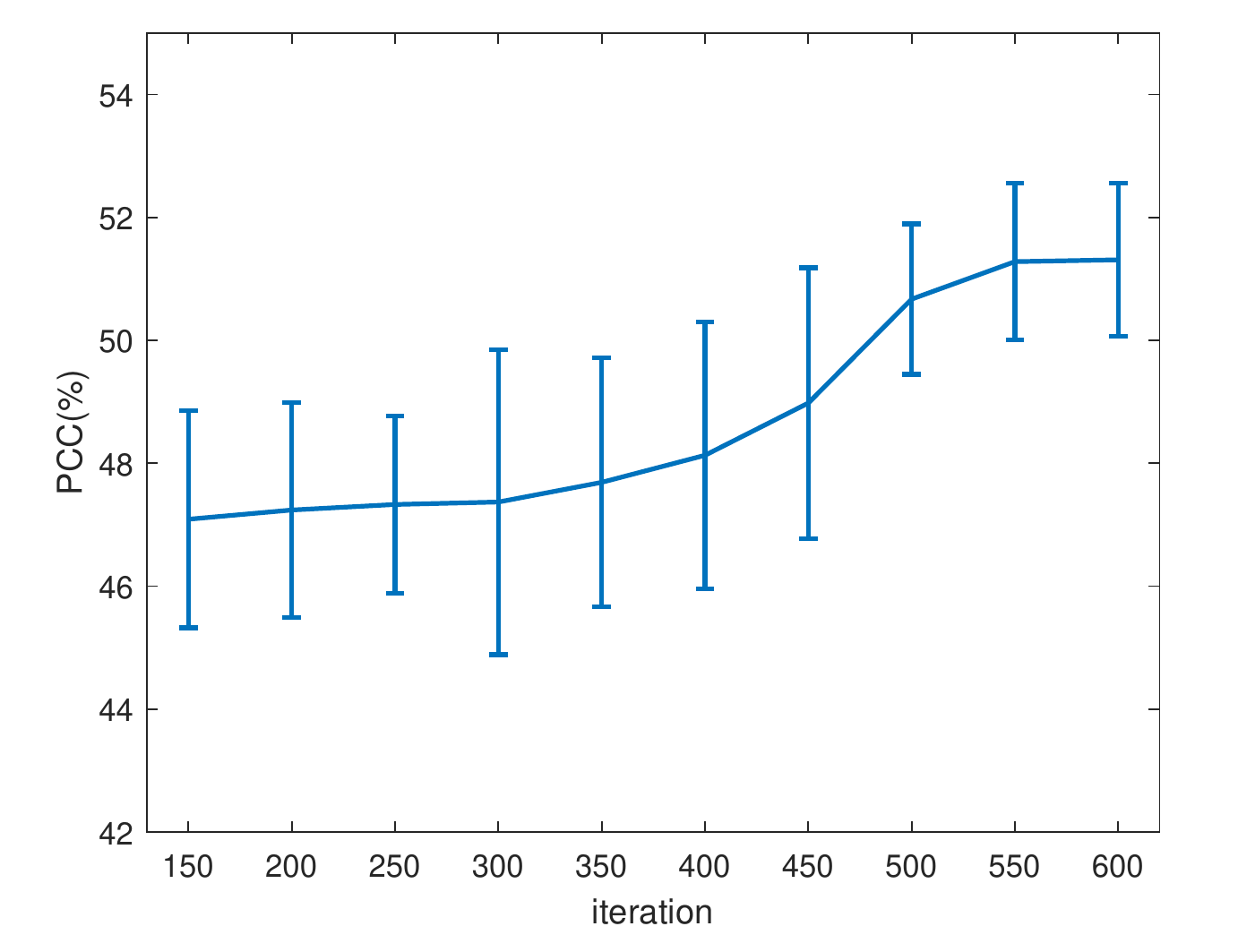}}
	\caption{The convergence analysis of the RMEN-CCA on MNIST and Wiki.}
	\label{fig:1}
\end{figure*}

In this section, we use the accuracy curve to illustrate the convergence analysis of the RMEN-CCA instead of the convergence curve on MNIST and Wiki, which is more visually pleasing for the convergence analysis. With the number of iterations increased, the accuracy of the RMEN-CCA is improved. When the accuracy curve is almost flat, we consider the RMEN-CCA almost converges. In Figure \ref{fig:1a}, we note that there is a jump in the curve, when the number of iterations is about 30, and the accuracy of the RMEN-CCA improves afterwards. When the iteration reaches 150, the RMEN-CCA almost converges. Therefore, we set the number of iteration as 150 on MNIST. Moreover, in Figure \ref{fig:1b}, we only illustrate an interval of the accuracy curve of the RMEN-CCA on Wiki because of the different dimensions of the two 'views', which is from 150 to 600. We note that the RMEN-CCA almost converges when the iteration reaches 500, and therewith the standard deviation decreases significantly. Therefore balancing the effectiveness and efficiency, we set the number of iteration as 500. Furthermore, we will investigate the convergence of the RMEN-CCA on the large-scale URL Reputation dataset, which is compared with the Scalable CCA. Due to space limitations, we omit the convergence analysis of the RMEN-CCA on Pascal VOC 2007.

\subsection{Performance}
\subsubsection{Results on MNIST}

In this paper, we only consider unsupervised learning approaches as comparisons, since this paper focuses on the multi-view unsupervised learning problems. The compared approaches include linear CCA, partial least squares (PLS) \cite{rosipal2006overview}, bilinear models (BLM) \cite{sharma2012generalized,tenenbaum1997separating}, FKCCA and NKCCA \cite{lopez2014randomized}, approximate KCCA (KNOI) \cite{wang2015large} as well as Scalable CCA \cite{ma2015finding}. To better illustrate the effectiveness of the RMEN, we also derive a CCA with the MEN method termed MEN-CCA. The KNOI is carried out on the GPU, while other algorithms are run on the CPU. The experimental results in terms of PCC and time are shown in Table \ref{table2}. 
\begin{table}[ht]
	\renewcommand{\arraystretch}{1.3}
	\caption{The comparison with state-of-the-art approaches on MNIST in terms of PCC (\%) and time (sec).}
	\label{table2}
	\centering
	\begin{tabular}{|c|c|c|}
		\hline
		Algorithms & PCC(\%)&time(sec) \\
		\hline
		RMEN-CCA & $\bm{93.11}$ &10.04\\
		\hline
		MEN-CCA & 90.82 & 8.19\\
		\hline
		KNOI & 87.26 & 258.52 (GPU) \\
		\hline
		FKCCA & 82.61 & 232.87 \\
		\hline
		NKCCA & 84.92 & 257.69 \\
		\hline
		Scalable CCA & 56.87 & 6.39\\
		\hline
		Linear CCA & 58.43 & 3.08 \\
		\hline
		PLS & 58.16 & 3.51\\
		\hline
		BLM & 58.68 & 2.89 \\
		\hline
	\end{tabular}
\end{table} 

Seen from Table \ref{table2}, the RMEN-CCA has the best performance among all comparisons. The RMEN-CCA outperforms remarkably Scalable CCA in terms of accuracy, since the RMEN-CCA benefits from the strength of coupled feature selection. The RMEN-CCA takes a little more time than Scalable CCA. Since the RMEN-CCA needs to calculate $\bm{S}$ using SVD, and the computational complexity is $O(n^3)$, in which $n$ is the number of input data. But the time cost of the RMEN-CCA is much less than that of all kernel-based methods, even if the KNOI is carried out on the GPU. We note that the above experimental results in terms of time may be unfair to the RMEN-CCA. Since all kernel-based algorithms are sensitive to the user-specified kernel width. As a result, many experiments on the validation set have been carried out for all kernel-based algorithms in order to achieve the reported results, and these additional time costs are not included in Table \ref{table2}.   

\subsubsection{Results on Wiki}

Due to the limitations of computational resource, we only evaluate the KRMEN-CCA on this dataset. In addition, the KCCA \cite{bach2002kernel} is also used as a comparison on this dataset. The experimental results are illustrated in Table \ref{table3}.

From Table \ref{table3}, except for the KRMEN-CCA and KCCA, the RMEN-CCA outperforms other comparisons. The KCCA sightly outperforms the RMEN-CCA, but the KCCA needs about 1,120 times as many time costs as the RMEN-CCA. Surprisingly, the KRMEN-CCA can achieve extremely great accuracy, while the time of the KRMEN-CCA is only one-third of that of the KCCA.
\begin{table}[ht]
	\renewcommand{\arraystretch}{1.3}
	\caption{The comparison with state-of-the-art approaches on Wiki in terms of PCC (\%) and time (sec).}
	\label{table3}
	\centering
	\begin{tabular}{|c|c|c|}
		\hline
		Algorithms & PCC(\%) & time(sec) \\
		\hline
		RMEN-CCA & 51.28 & 1.27 \\
		\hline
		MEN-CCA & 50.33 & 1.09 \\
		\hline
		KRMEN-CCA  & $\bm{95.23}$ & 465.71 \\
		\hline
		KCCA & 58.54 & 1427.67 \\
		\hline
		KNOI & 49.61 & 213.45 \\
		\hline 
		FKCCA & 50.12 & 143.61\\
		\hline
		NKCCA & 50.64 & 159.28\\
		\hline
		Scalable CCA & 46.19 & 0.43 \\
		\hline
		Linear CCA & 46.36 & 0.07 \\
		\hline
		PLS & 46.32 & 0.08\\
		\hline
		BLM & 46.92 & 0.07 \\
		\hline
	\end{tabular}
\end{table}
\subsubsection{Results on Pascal VOC 2007}
In order to illustrate better the RMEN-CCA in terms of effectiveness and efficiency, we not only evaluate the RMEN-CCA against some state-of-the-art algorithms on image-to-tags features based Pascal VOC 2007, but also conduct an additional group of experiments on image-to-image features based Pascal VOC 2007.

As shown by both Table \ref{table4} and \ref{table5}, we see that the RMEN-CCA outperforms all comparisons in terms of accuracy at much faster learning speed. 

\begin{table*}[ht]
	\renewcommand{\arraystretch}{1.3}
	\caption{The comparison with state-of-the-art approaches on image-to-tags features based Pascal VOC 2007 in terms of PCC (\%) and time (sec).}
	\label{table4}
	\centering
	\begin{tabular}{|c|c|c|c|c|c|c|}
		\hline
		\multirow{2}{*}{Algorithms} 
		& \multicolumn{2}{c|}{Gist-Wc} & \multicolumn{2}{c|}{Gist-Rel} & \multicolumn{2}{c|}{Gist-Abs} \\
		\cline{2-7}
		& PCC(\%) & time(sec)& PCC(\%) & time(sec)& PCC(\%) & time(sec)\\
		\hline
		RMEN-CCA & $\bm{65.88}$ & 0.89 & $\bm{67.19}$ & 0.96 &  $\bm{59.53}$ & 0.95\\
		\hline
		MEN-CCA & 60.29 & 0.71 & 63.18 & 0.74 &  57.10 & 0.82\\
		\hline
		KNOI & 54.51 & 522.88 &  57.63 & 516.79 & 56.21& 523.24\\
		\hline
		FKCCA & 55.69 & 157.18 & 55.41 & 153.84 & 56.63 & 154.68\\
		\hline
		NKCCA & 55.63 & 163.24 & 55.77 & 143.64 & 57.29 & 171.46 \\
		\hline
		Scalable CCA & 51.53 & 0.88 & 51.36 & 0.84 & 53.56& 0.79\\
		\hline
		Linear CCA & 51.77 & 0.87 & 52.40 & 0.81 & 53.63& 0.81\\
		\hline
		PLS & 51.75 & 0.82 & 51.65 & 0.83 & 53.27& 0.80\\
		\hline
		BLM & 52.08 & 0.92 & 51.94 & 0.85 & 53.91 & 0.89\\
		\hline
	\end{tabular}
\end{table*}
\begin{table*}[ht]
	\renewcommand{\arraystretch}{1.3}
	\caption{The comparison with state-of-the-art approaches on image-to-image features based Pascal VOC 2007 in terms of PCC (\%) and time (sec).}
	\label{table5}
	\centering
	\begin{tabular}{|c|c|c|c|c|c|c|}
		\hline
		\multirow{2}{*}{Algorithms} 
		& \multicolumn{2}{c|}{Gist-Bow} & \multicolumn{2}{c|}{Gist-Hsv}& \multicolumn{2}{c|}{Hsv-Bow}  \\
		\cline{2-7}
		& PCC(\%) & time(sec)& PCC(\%) & time(sec)& PCC(\%) & time(sec)\\
		\hline
		RMEN-CCA & $\bm{66.35}$ & 0.84 & $\bm{62.16}$ & 0.81 & $\bm{52.78}$ & 0.81 \\
		\hline
		MEN-CCA & 61.41 & 0.74 & 56.82 & 0.69 &  50.83 & 0.68\\
		\hline
		KNOI & 61.54 & 333.22 & 54.27 & 392.74 & 47.66 & 375.42\\
		\hline
		FKCCA & 61.60 & 156.62 & 50.83 & 148.60 & 50.76 & 164.53\\
		\hline
		NKCCA & 61.83 & 147.27 & 53.19 & 166.24 & 50.69 & 187.27\\
		\hline
		Scale CCA & 47.99 & 0.82 & 49.76 & 0.71 & 41.91 & 0.67 \\
		\hline
		Linear CCA & 49.63  & 0.72 & 50.18 & 0.68 & 43.48 & 0.56\\
		\hline
		PLS & 48.23 & 0.79 & 50.05 & 0.69 & 43.17& 0.62\\
		\hline
		BLM & 48.99 & 0.72 & 51.33 & 0.67 & 44.08 & 0.58\\
		\hline
	\end{tabular}
\end{table*}  

\subsubsection{Results on URL Reputation}
We only compare with Scalable CCA on the large-scale URL Reputation dataset. the previous work \cite{ma2015finding} has shown that the Scalable CCA can achieve excellent performance on this dataset, and thus, we omit other methods in order to avoid duplication of work. Classical CCA methods based on eigenvector computation fail on a typical PC machine. Since these approaches are trained over the entire training set in a batch learning fashion. Consequently, these methods are prohibitive for huge datasets. However, likewise to the Scalable CCA, the RMEN-CCA is also an online learning algorithm, which is a commonly-used approach on huge datasets.

\begin{table}[ht]
	\renewcommand{\arraystretch}{1.3}
	\caption{The comparison with Scalable CCA on the large-scale URL Reputation dataset in terms of PCC (\%) and time (sec).}
	\label{table6}
	\centering
	\begin{tabular}{|c|c|c|c|}
		\hline
		Algorithms & iteration & PCC(\%) & time(sec) \\
		\hline
		RMEN-CCA & 200 & $\bm{44.107}$ & 22.04 \\
		\hline
		\multirow{2}{*} {Scalable CCA}  
		& 200 & 35.814 & 21.93 \\
		\cline{2-4}
		& 1000 & 41.144 & 31.44 \\
		\hline
	\end{tabular}
\end{table}

\begin{figure}[ht]
	\renewcommand{\arraystretch}{1.3}
	\centering
	\includegraphics[width=3.3in]{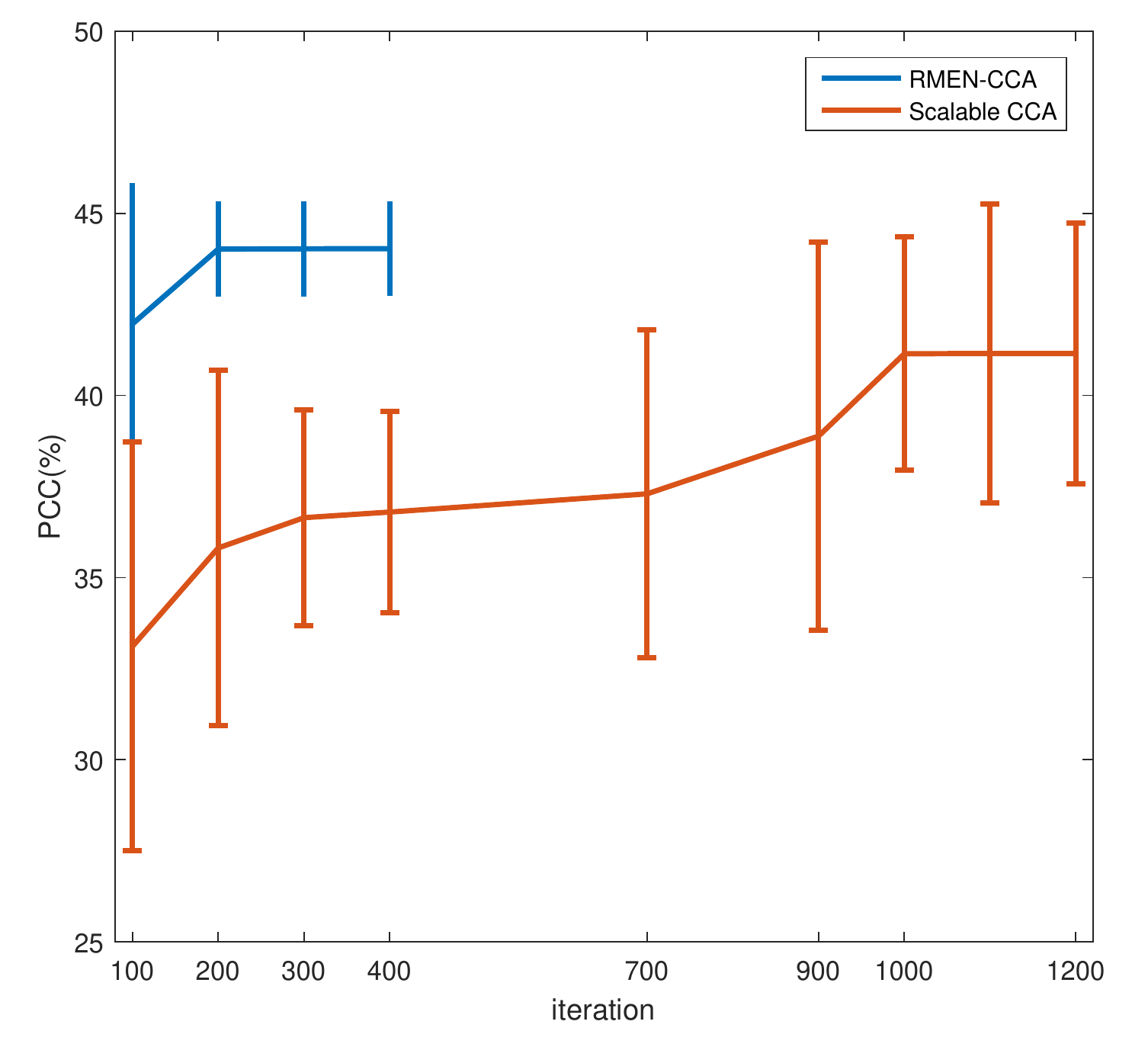}
	\caption{The convergence analysis of the RMEN-CCA and Scalable CCA on URL Reputation.}\label{fig:2}
\end{figure}

From Table \ref{table6}, we see that the RMEN-CCA can achieve better performance than the Scalable CCA. As shown by Figure \ref{fig:2}, the more the number of iteration is, the more the Scalable CCA can capture correlations. However, we find that  when the number of  iterations is more than 1,000, the Scalable CCA is not powerful enough to achieve better accuracy, that is, the PCC of the Scalable CCA remains about $41\%$. While the RMEN-CCA significantly outperforms the Scalable CCA no matter what the number of iterations of the Scalable CCA is. The experimental results on this large-scale dataset illustrate that the RMEN-CCA not only faster converges than the Scalable CCA, but this novel algorithm is also more stable than the Scalable CCA.

\section{Conclusion}
\label{sec:6}

In this paper, we derive a novel robust matrix elastic net based canonical correlation analysis (RMEN-CCA) with theoretically guaranteed convergence and empirical proficiency. To the best of our knowledge, the RMEN-CCA is the first to impose coupled feature selection into the CCA. As a consequence, the RMEN-CCA not only measures correlations between different 'views', but also distills numerous relevant and useful features, which results that the performance of the RMEN-CCA is significantly improved. Additionally, for the sake of modeling highly sophisticated nonlinear relationships, the RMEN-CCA can be extended straightforwardly to the kernel scenario. Furthermore, due to its complicated model architecture, it is nontrivial to solve the RMEN-CCA by existing optimization approaches. Therefore, we bridge the gap between the new optimization problem and the previous efficient iterative algorithm. Finally, competitive experimental results on four popular datasets confirm the great effectiveness and efficiency of the RMEN-CCA in multi-view unsupervised learning problems.

\section*{Acknowledgment}

The authors would like to thank Prof. Daniel Palomar from Hong Kong University of Science and Technology for great inspiration and valuable discussions of this work. The authors also would like to thank Prof. Ajay Joneja from Hong Kong University of Science and Technology for providing a PC machine and some constructive suggestions on this research.

\ifCLASSOPTIONcaptionsoff
  \newpage
\fi



%
\bibliographystyle{IEEEtran}
\bibliography{cca.bib}

%




\end{document}